%% file: main_arxiv.tex
\documentclass[11pt,a4paper]{article}

\usepackage{times,latexsym}
\usepackage[T1]{fontenc}
\usepackage{url}
\usepackage{amsmath,amssymb,amsthm,mathtools}
\usepackage{graphicx}
\usepackage{microtype}
\usepackage{booktabs}
\usepackage{multirow}
\usepackage[table]{xcolor}
\usepackage{siunitx}
\usepackage{placeins}
\usepackage{float}
\usepackage{enumitem}
\usepackage{xspace}
\usepackage{tabularx}
\usepackage{array}
\usepackage{listings}
\usepackage[acceptedWithA]{tacl2021v1}

\definecolor{BackboneGray}{HTML}{F2F2F2}
\definecolor{OursBlue}{HTML}{EAF2FB}

\definecolor{GainOne}{HTML}{EAF3F8}
\definecolor{GainTwo}{HTML}{D7E9F3}
\definecolor{GainThree}{HTML}{C2DDEA}

\definecolor{LossOne}{HTML}{FAEEEE}
\definecolor{LossTwo}{HTML}{F4DDDC}
\definecolor{LossThree}{HTML}{EEC7C4}

\definecolor{CrossLinkRed}{HTML}{7A4141}
\definecolor{TakeawayBorder}{HTML}{9A6661}
\definecolor{TakeawayBg}{HTML}{F8F2F1}

\definecolor{darkblue}{rgb}{0,0,0.5}

\definecolor{PromptBg}{HTML}{F7F7F7}
\definecolor{PromptBorder}{HTML}{B9B9B9}

\hypersetup{
    colorlinks=true,
    citecolor=darkblue,
    linkcolor=CrossLinkRed,
    urlcolor=darkblue
}

\newcommand{\method}{\textsc{BeliefRAG}\xspace}

\newcommand{\gainone}[1]{\cellcolor{GainOne}#1}
\newcommand{\gaintwo}[1]{\cellcolor{GainTwo}#1}
\newcommand{\gainthree}[1]{\cellcolor{GainThree}#1}

\newcommand{\lossone}[1]{\cellcolor{LossOne}#1}
\newcommand{\losstwo}[1]{\cellcolor{LossTwo}#1}
\newcommand{\lossthree}[1]{\cellcolor{LossThree}#1}

\lstnewenvironment{promptblock}[1][]
{
\lstset{
    basicstyle=\ttfamily\fontsize{11}{11.3}\selectfont,
    backgroundcolor=\color{PromptBg},
    rulecolor=\color{PromptBorder},
    frame=single,
    framesep=2pt,
    xleftmargin=3pt,
    xrightmargin=3pt,
    aboveskip=3pt,
    belowskip=3pt,
    breaklines=true,
    breakatwhitespace=false,
    columns=fullflexible,
    keepspaces=true,
    showstringspaces=false,
    tabsize=2,
    #1
}}
{}

\newcommand{\keytakeaway}[1]{%
    \par\vspace{1pt}\noindent
    {%
    \setlength{\fboxsep}{3pt}%
    \fcolorbox{TakeawayBorder}{TakeawayBg}{%
        \parbox{%
            \dimexpr\linewidth-2\fboxsep-2\fboxrule\relax
        }{%
            \textbf{Key takeaway.} #1
        }%
    }%
    }%
    \par\vspace{1pt}
}


\title{
    \method:
    Making Adaptive RAG State-Aware under Evolving Evidence
}

\author{
    Hongji Pu \\
    University of Illinois Urbana-Champaign \\
    \texttt{hongjip2@illinois.edu}
}

\date{}

\begin{document}

\maketitle


\begin{abstract}
Adaptive RAG uses signals such as confidence, relevance, support, and
retrieval quality to decide when to search or correct evidence.
In multi-step retrieval, however, these local signals must be combined
into a persistent view of what the current evidence supports, what
remains missing, and which action should follow.
Existing methods often use such signals as separate triggers, making
it difficult to preserve a coherent evidence state across a trajectory;
we call this problem \emph{evidence-state fragmentation}.
We introduce \textsc{BeliefRAG}, a closed-loop controller that updates
an explicit state over sufficiency, reliability, conflict, uncertainty,
evidence gaps, and acquisition cost, then chooses among retrieval,
query rewriting, verification, answering, stopping, and abstention.
Across six QA benchmarks with gpt-oss-120b,
\textsc{BeliefRAG} reaches mean token F1 $0.572$ with $3.89$k tokens
per question, outperforming fixed iterative retrieval ($0.555$ F1)
while using $39\%$ fewer tokens.
The same quality--cost pattern transfers to Qwen3-32B, where
\textsc{BeliefRAG} reaches $0.552$ F1 versus $0.523$ for iterative
retrieval while using $35\%$ fewer tokens.
Analysis shows that the main gains come from corrective re-retrieval
rather than pruning alone, while several belief dimensions are
redundant and calibrated answerability plays the strongest operational
role.
Calibration improves threshold stability across related evidence
sources, although source shift can still invalidate the same decision
signal.
\end{abstract}


\input{sections/1_Introduction}

\input{sections/2_Related_Work}

\input{sections/3_Methodology}

\input{sections/4_experiments}

\input{sections/5_result}

\input{sections/6_Analysis_Conclusion}


\clearpage

\bibliographystyle{acl_natbib}
\bibliography{reference}


\clearpage
\appendix

\input{sections/Appendix}

\end{document}

%% file: sections/1_Introduction.tex
\section{Introduction}

Retrieval-augmented generation (RAG) equips language models with external
evidence that can supplement knowledge stored in model parameters
\citep{lewis2020retrieval}.
The standard RAG pipeline retrieves a fixed set of passages once and generates
an answer from the retrieved context \citep{lewis2020retrieval}.
Multi-hop questions often require several pieces of evidence that become
available only after intermediate entities or facts have been identified
\citep{trivedi2022musique}.
A retrieval system for these questions therefore needs to decide repeatedly
what information is still missing, whether another search is useful, and when
the collected evidence is sufficient for answering.

Adaptive RAG introduces such decisions into the retrieval process.
FLARE triggers retrieval from low confidence during generation
\citep{jiang2023active}.
Self-RAG learns reflection signals for retrieval need, evidence relevance,
answer support, and response utility \citep{asai2024selfrag}.
Adaptive-RAG predicts question complexity and selects no retrieval, one-step
retrieval, or iterative retrieval \citep{jeong2024adaptiverag}.
DRAGIN estimates the information currently needed by the model and uses that
estimate to determine retrieval timing and queries \citep{su2024dragin}.
CRAG evaluates retrieved evidence and invokes corrective retrieval when the
initial evidence is judged poor \citep{yan2024corrective}.
Together, these methods establish confidence, relevance, support, information
need, and retrieval quality as useful signals for adaptive retrieval.

A shared problem appears once retrieval lasts for several steps:
\emph{the controller must convert several local signals into one decision about
the current evidence}.
The same low-confidence state can arise because a required fact is missing,
because the retained passages disagree, or because the available evidence is
too weak to support an answer.
These situations require different actions.
Missing information motivates further retrieval or query rewriting.
Conflicting evidence motivates verification.
Sufficient evidence motivates termination.
Low-value future retrieval motivates stopping even when the evidence remains
imperfect.
Partially observable decision problems commonly address this type of setting by
maintaining a belief over information that cannot be observed directly
\citep{kaelbling1998planning}.

The difficulty comes from the meaning and interaction of the available
signals.
Self-RAG shows that relevance and support provide distinct judgments about
retrieved evidence \citep{asai2024selfrag}.
DRAGIN shows that retrieval can be driven by the model's current information
need \citep{su2024dragin}.
Astute RAG studies unreliable evidence and conflicts between retrieved and
internal knowledge \citep{wang2025astute}.
Adaptive-RAG demonstrates that the useful amount of retrieval varies across
questions \citep{jeong2024adaptiverag}.
These signals describe different aspects of one evolving evidence condition.
Their values can also be redundant, poorly calibrated, or weakly connected to
the controller's actual actions.
A useful diagnostic therefore requires both informative measurements and a
decision rule that can consume those measurements at values reached during
real trajectories.

We call this problem \textbf{evidence-state fragmentation}.
A retrieval trajectory contains queries, passages, verification outcomes, and
previous actions.
The controller needs a compact summary of what these observations currently
imply about the evidence.
BeliefRAG provides this summary as a persistent evidence state.
At each step, it measures relevance, support, conflict, uncertainty, remaining
evidence gaps, novelty, and acquisition cost.
These observations update six operational belief dimensions:
sufficiency, reliability, conflict, uncertainty, evidence gap, and cost.
The six dimensions form an explicit design hypothesis.
Our experiments test their incremental information and their actual influence
on controller actions.

\begin{figure*}[t]
    \centering
    \includegraphics[width=\textwidth]{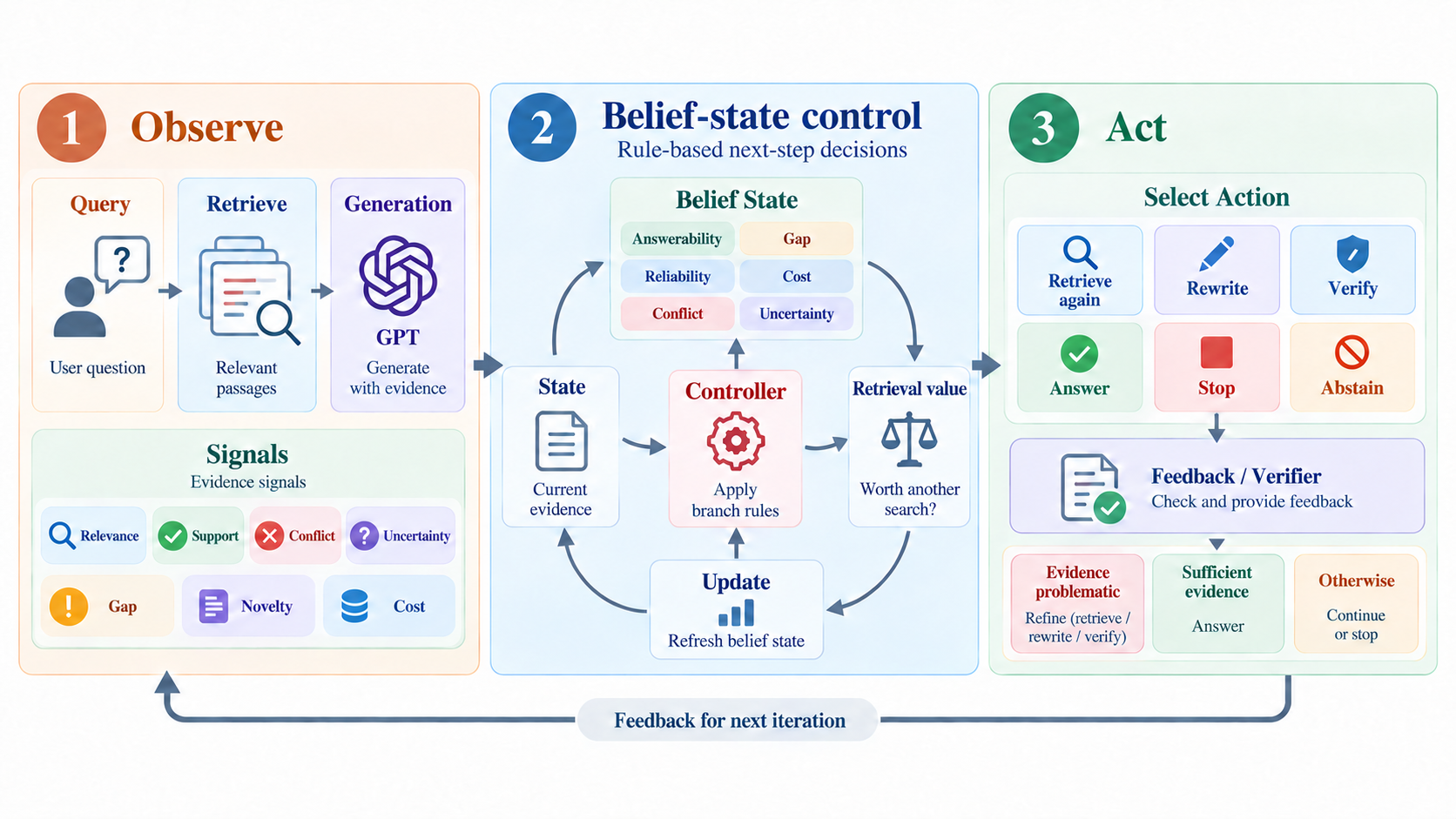}
\caption{
\textbf{BeliefRAG as a closed-loop evidence controller.}
The system converts the current evidence into diagnostic signals and a compact
belief state, then applies fixed branch rules to choose among retrieval,
rewriting, verification, answering, stopping, and abstention. Retrieval value
and verifier feedback determine whether more evidence is worth acquiring and
update the next iteration. The controller is non-RL; only the sufficiency and
retrieval-value estimates are fitted from data.
}
    \label{fig:bayes_rl_rag_framework}
\end{figure*}

The updated state controls six actions:
\textsc{Retrieve}, \textsc{Rewrite}, \textsc{Verify}, \textsc{Answer},
\textsc{Stop}, and \textsc{Abstain}.
Evidence with material conflict or low reliability can enter a corrective
loop that verifies, removes weak passages, rewrites the query, and retrieves
replacement evidence.
Evidence with high answerability can terminate with an answer.
Insufficient evidence can trigger another retrieval when the estimated value
of another search is high enough.
Low novelty can trigger query rewriting.
Low expected acquisition value can terminate further search.
Figure~\ref{fig:bayes_rl_rag_framework} summarizes this closed-loop process.

We evaluate BeliefRAG under a controlled protocol in which compared methods share the same corpus, retriever, verifier, budget, and evaluation examples within each backbone. The primary table uses gpt-oss-120b: BeliefRAG reaches mean token F1 $0.572$ with $3.89$k tokens per question, outperforming fixed Iterative RAG at $0.555$ F1 while using $39\%$ fewer tokens. We repeat the six-dataset evaluation with Qwen3-32B in Appendix Tables~\ref{tab:qwen_answer_metrics}--\ref{tab:qwen_cost_recall}; there BeliefRAG reaches $0.552$ mean F1 versus $0.523$ for Iterative RAG with $35\%$ fewer tokens. The analyses show that corrective re-retrieval, calibrated answerability, and stable decision thresholds explain the quality--cost trade-off, while evidence-source shift remains a clear boundary.

Our contributions are threefold:
\begin{itemize}[leftmargin=*,topsep=2pt,itemsep=1pt,parsep=0pt,partopsep=0pt]
    \item \textbf{Problem.} We identify \textbf{evidence-state fragmentation}: multi-step RAG lacks a persistent state that summarizes what multiple evidence signals imply for the next action.
    
    \item \textbf{Method.} We propose \textbf{BeliefRAG}, a closed-loop controller that updates an explicit evidence state and uses it to coordinate retrieval, rewriting, verification, answering, stopping, and abstention.
    
    \item \textbf{Findings.} Controlled experiments across two backbones show a strong quality--cost trade-off and reveal which mechanisms matter in practice: corrective re-retrieval, calibrated answerability, and transferable decision thresholds.
\end{itemize}

%% file: sections/2_Related_Work.tex
\section{Related Work}

\textbf{Adaptive RAG.}
RAG retrieves evidence once before generation, leaving search depth fixed across queries~\citep{lewis2020retrieval}.
Self-RAG learns reflection tokens for retrieval, relevance, support, and utility, but these judgments remain local to the current step~\citep{asai2024selfrag}.
CRAG uses a retrieval-quality evaluator to trigger correction when evidence is poor, making evaluator reliability a key failure point~\citep{yan2024corrective}.
Adaptive-RAG routes queries among no retrieval, one-shot retrieval, and iterative retrieval, trading task difficulty against unnecessary search cost~\citep{jeong2024adaptiverag}.

 \textbf{Uncertainty and adaptive retrieval.}
Adaptive retrieval must decide whether more evidence is still useful.
Uncertainty is a natural signal, but its reliability varies across tasks
and estimators~\citep{moskvoretskii2025adaptive}.
In RAG, retrieved evidence can also change the meaning of confidence, so
standard uncertainty scores may no longer track answer correctness
\citep{soudani2025uncertainty}.
SeaKR uses internal model states to guide retrieval and reranking
\citep{yao2025seakr}, while CtrlA learns control signals directly from
representations~\citep{huanshuo2025ctrla}.
These results suggest that no single confidence signal is uniformly reliable.

\textbf{Evidence quality and conflict.}
Topically relevant evidence can still be misleading, and models may
over-weight relevance when judging its quality \citep{wan2024evidence}.
Retrieved evidence can conflict with parametric knowledge, biasing models
toward faulty internal memory \citep{jin2024tug}.
Conflicting sources can also substantially degrade RAG performance
\citep{pham2024whoswho}.
Retrieved context can make standard uncertainty estimates unreliable
\citep{soudani2025uncertainty}.
BeliefRAG therefore tracks evidence quality, conflict, and answerability
across steps rather than relying on one relevance or confidence score.

%% file: sections/3_Methodology.tex
\section{Methodology}
\label{sec:method}

\subsection{Overview}

BeliefRAG controls \emph{what to do after each retrieval}.
A retrieved passage being relevant does not yet mean that the question can be
answered: a required fact may still be missing, two passages may disagree, or
another search may simply repeat what is already known.
The controller therefore makes three decisions from the evidence accumulated
so far:
\emph{should the evidence be corrected, is it sufficient to answer, and if not,
is another retrieval worth doing?}

At step $t$, the agent maintains
\begin{equation}
s_t=(q,E_t,H_t,b_t),
\label{eq:state}
\end{equation}
where $q$ is the question, $E_t$ is the evidence currently retained,
$H_t$ records previous queries and actions, and $b_t$ summarizes the current
evidence condition.
Each step follows
\[
\begin{aligned}
\text{measure }E_t &\rightarrow \text{update }b_t \\
&\rightarrow \text{choose an action} \rightarrow \text{update }E_t .
\end{aligned}
\]
Importantly, $E_t$ is not an append-only retrieval history:
\textsc{Retrieve} can add passages, while \textsc{Verify} can remove passages
judged unhelpful.

\FloatBarrier
\subsection{Measuring the Current Evidence}

Before making a decision, BeliefRAG computes seven diagnostics,
\begin{equation}
x_t=[R_t,S_t,C_t,U_t,G_t,N_t,K_t].
\label{eq:diagnostics}
\end{equation}
They answer concrete questions about the evidence and are computed as shown
in Table~\ref{tab:diagnostics}.

\begin{table}[t]
\centering
\caption{Evidence diagnostics used at each decision step.}
\label{tab:diagnostics}
\scriptsize
\setlength{\tabcolsep}{2.2pt}
\renewcommand{\arraystretch}{1.04}
\begin{tabular}{@{}c l p{0.58\columnwidth}@{}}
\toprule
 & \textbf{Meaning} & \textbf{Computation} \\
\midrule
$R_t$ & Relevance &
Softmax-weighted mean of calibrated top-$k$ retrieval scores. \\

$S_t$ & Support &
Verifier score for how strongly $E_t$ supports a complete answer. \\

$C_t$ & Conflict &
Largest material contradiction within $E_t$ or between evidence and the
current draft. \\

$U_t$ & Uncertainty &
Verifier estimate of how uncertain the answer remains given only $E_t$. \\

$G_t$ & Gap &
Estimated fraction of information required by the question that is still
unsupported. \\

$N_t$ & Novelty &
$1-\mathrm{sim}(\Delta E_t,E_{t-1})$; high when the newest retrieval adds
information not already retained. \\

$K_t$ & Cost &
$\min(1,\text{tokens used}/\text{token budget})$. \\
\bottomrule
\end{tabular}
\end{table}

The four semantic quantities $S_t,C_t,U_t,G_t$ are produced together by one
structured verifier call.
$R_t$, $N_t$, and $K_t$ are computed locally.
For example, the raw retriever score $z_{t,i}$ of passage $i$ at step $t$ is first mapped to $[0,1]$ by
\[
\tilde z_{t,i}
=
\sigma\!\left((z_{t,i}-\mu_s)/\tau_s\right),
\]
where $\sigma(z)=1/(1+e^{-z})$ is the logistic sigmoid, $\mu_s$ is the
retriever-score location statistic, and $\tau_s>0$ is its scale. $R_t$ is then
the softmax-weighted mean of the top-$k$ calibrated scores, where $k$ is the number of passages returned by one retrieval call.
In Table~\ref{tab:diagnostics}, $\Delta E_t$ denotes passages newly returned at
step $t$, and $\mathrm{sim}$ is the configured passage-similarity function.
Novelty compares these new passages with those already retained.
Cost is not predicted by the model: it is the fraction of the token budget
already consumed.
Thus retrieval rounds, top-$k$, action count, and token budget remain
different constraints rather than one vague ``capacity'' variable.

\subsection{From Diagnostics to Belief}

The diagnostics describe individual properties of $E_t$.
BeliefRAG converts them into six quantities with direct decision meanings:
\begin{equation}
b_t=[
b_t^{\mathrm{suff}},
b_t^{\mathrm{rel}},
b_t^{\mathrm{conf}},
b_t^{\mathrm{unc}},
b_t^{\mathrm{gap}},
b_t^{\mathrm{cost}}
].
\label{eq:belief}
\end{equation}

\textbf{Sufficiency} means that the retained evidence is enough to answer;
\textbf{reliability} means that the retained sources appear trustworthy;
\textbf{conflict} means that the evidence materially disagrees;
\textbf{uncertainty} means that the answer is still unclear;
\textbf{gap} means that required information is still missing; and
\textbf{cost} records how much acquisition budget has already been spent.

These quantities are related but not interchangeable.
For example, a passage may be highly relevant and well supported but still
leave the second hop of a multi-hop question unresolved.
Similarly, two individually relevant passages may contradict each other.
Sufficiency therefore asks a higher-level question than relevance or support:
\emph{can the question be answered correctly from the evidence retained now?}

For each inferred dimension, the complete diagnostic vector is mapped to an
instantaneous belief. Here $\alpha_d$ is a dimension-specific intercept,
$w_d$ is its diagnostic-weight vector, and $d$ indexes the five inferred
(non-cost) dimensions:
\begin{equation}
\begin{aligned}
\hat b_t^d &= \sigma(\alpha_d+w_d^\top x_t),\\[-2pt]
d &\in \{\mathrm{suff},\mathrm{rel},\mathrm{conf},
\mathrm{unc},\mathrm{gap}\}.
\end{aligned}
\label{eq:belief_est}
\end{equation}
For example, sufficiency increases with support and relevance and decreases
with evidence gap, uncertainty, and conflict; reliability increases with
relevance and support but decreases with conflict.
Cost is observed directly:
$b_t^{\mathrm{cost}}=K_t$.

Because one verifier call can be noisy, the new estimate is blended with the
previous belief, where $\lambda_d\in[0,1]$ is the weight placed on the current
observation in logit space:
\begin{equation}
\begin{aligned}
b_t^d=\sigma\big(
 &(1-\lambda_d)\operatorname{logit}(b_{t-1}^d)\\
 &+\lambda_d\operatorname{logit}(\hat b_t^d)
\big).
\end{aligned}
\label{eq:belief_update}
\end{equation}
The exact coefficients and initial values are reported in
Appendix~\ref{app:implementation}.

\paragraph{Operational answerability.}
The persistent belief coordinate $b_t^{\mathrm{suff}}$ and the answer gate are distinct. The former summarizes evidence sufficiency; the latter uses
\begin{equation}
p_t^{\mathrm{ans}}=P(\mathrm{answerable}\mid q,E_t),
\label{eq:pans}
\end{equation}
where ``answerable'' means that the frozen generator produces a correct answer under the fixed prompt. Because that prompt requests a best short answer even when documents are incomplete, $p_t^{\mathrm{ans}}$ may reflect frozen parametric knowledge and is not evidence-only entailment. We fit $p_t^{\mathrm{ans}}=\sigma(\alpha_{\mathrm{ans}}+w_{\mathrm{ans}}^\top x_t)$ on HotpotQA train, select it on development data, and freeze it across all six benchmarks; Appendix~\ref{app:answerability_calibration} gives the fitted parameters.

\FloatBarrier
\subsection{How Beliefs Produce Actions}

BeliefRAG chooses among six actions:
\[
\begin{aligned}
\mathcal{A}=\{&\textsc{Retrieve},\textsc{Rewrite},\textsc{Verify},\\[-2pt]
&\textsc{Answer},\textsc{Stop},\textsc{Abstain}\}.
\end{aligned}
\]
Rather than scoring them as unrelated choices, the main controller evaluates
a short sequence of questions shown in Table~\ref{tab:policy}.

\begin{table}[t]
\centering
\caption{Main decision rules. Values are the default operating thresholds;
experiment-specific selected values are reported in the Appendix.}
\label{tab:policy}
\scriptsize
\setlength{\tabcolsep}{2.4pt}
\renewcommand{\arraystretch}{1.06}
\begin{tabular}{@{}p{0.21\columnwidth}p{0.47\columnwidth}p{0.22\columnwidth}@{}}
\toprule
\textbf{Decision} & \textbf{Condition} & \textbf{Default} \\
\midrule

Correct &
Evidence is materially conflicting or clearly unreliable &
$b_t^{\rm conf}\ge0.50$ or
$b_t^{\rm rel}<0.30$ \\

Answer &
Current state is operationally answerable and conflict is acceptable &
$p_t^{\rm ans}\ge0.50$ \\

Retrieve &
Evidence is insufficient, budget remains, and another retrieval has enough
chance to make it answerable &
$p_t^{\rm flip}\ge0.10$ \\

Rewrite &
Another retrieval is useful, but the previous retrieval added little new
information &
$N_t<0.20$ \\

Stop / Abstain &
Evidence is still insufficient and further acquisition has low expected value &
no useful acquisition \\
\bottomrule
\end{tabular}
\end{table}

\paragraph{Correcting bad evidence.}
The controller checks correction before answering.
A high conflict score means that the retained passages materially disagree;
low reliability means that their combined relevance/support pattern is not
trustworthy enough.
The verifier can also return identifiers of passages it considers unhelpful;
when that trigger is enabled, those passages provide an additional correction
signal.

Correction is not simply deletion:
\[
\textsc{Verify}
\rightarrow
\text{prune}
\rightarrow
\textsc{Rewrite}
\rightarrow
\textsc{Retrieve}.
\]
Verification first removes the problematic passages, then the query is
reformulated and a replacement retrieval is issued.
This design matters because deleting weak evidence without replacing the
missing information may leave the question even less answerable.

\paragraph{Deciding when to answer.}
After correction is considered, the controller checks
$p_t^{\mathrm{ans}}$.
The default threshold is $\tau_{\mathrm{ans}}=0.50$: under the probability
interpretation, the current evidence must be at least as likely to be
answerable as not.
The threshold is an operating point rather than a universal constant; when it
is selected on a development/selection split, it is frozen before final
evaluation.

\paragraph{Deciding whether to retrieve again.}
If $p_t^{\mathrm{ans}}<\tau_{\mathrm{ans}}$, the agent does not automatically
retrieve merely because it is uncertain.
It asks a second question:
\emph{is one more retrieval likely to change the state from insufficient to
sufficient?}
We estimate
\begin{equation}
p_t^{\mathrm{flip}}
=
\sigma\!\left(
\alpha_{\mathrm{flip}}+w_r^{\mathrm{flip}} r_t+w_N^{\mathrm{flip}} N_t+w_s^{\mathrm{flip}} b_t^{\mathrm{suff}}
\right),
\label{eq:pflip}
\end{equation}
where $r_t$ is the number of retrieval rounds already used and
$\alpha_{\mathrm{flip}},w_r^{\mathrm{flip}},w_N^{\mathrm{flip}},w_s^{\mathrm{flip}}$ are retrieval-value coefficients fitted on the fit split
(or replaced by the fixed fallback schedule reported in the Appendix).
The default minimum acquisition value is $0.10$.
Thus another search is attempted only when it has at least the required
estimated chance of making the evidence answerable and retrieval budget
remains.

Novelty then decides \emph{how} to continue.
If the previous retrieval added useful new information
($N_t\ge0.20$ by default), the controller can issue another retrieval.
If novelty is below $0.20$, repeating essentially the same query is unlikely
to help, so the controller prefers \textsc{Rewrite} before searching again.

\paragraph{Stopping and abstention.}
If the evidence is not sufficiently answerable and another acquisition has
low value, the controller stops spending the remaining budget.
\textsc{Stop} means ``use the evidence we have and produce the best answer'';
\textsc{Abstain} means ``the evidence is inadequate and no useful acquisition
remains.''
Low confidence alone is therefore not an abstention rule:
uncertainty must be combined with evidence insufficiency and low acquisition
value.

\subsection{Controlled Evaluation}

The retriever, generator, verifier, answer prompt, evaluator, and budget are
held fixed across compared controllers.
The main answerability calibrator is fitted once on HotpotQA train and
selected on HotpotQA development data, then reused unchanged across all six
benchmarks. Separate fit/selection/holdout partitions are used for the causal
analyses. Exact thresholds, fitted coefficients, prompts, budget limits, and
reproduction details are given in Appendix~\ref{app:implementation}.

%% file: sections/4_experiments.tex
\section{Experiments}
\label{sec:experiments}

We evaluate \method in one shared harness so that differences come from the retrieval controller rather than from different tools or prompts. The language model acts as the agent's generator: it writes search queries and final answers, but it cannot retrieve documents or execute actions by itself. The harness executes every \textsc{Retrieve}, \textsc{Rewrite}, \textsc{Verify}, \textsc{Answer}, \textsc{Stop}, or \textsc{Abstain} action and records the resulting evidence, belief state, and cost. All compared methods therefore share the same generator, retriever, verifier, answer prompt, evaluator, and budget.

\subsection{Tasks and Benchmarks}

The main experiment uses six QA benchmarks with two different roles. HotpotQA~\citep{yang2018hotpotqa}, 2WikiMultiHopQA~\citep{ho2020wikimultihop}, and MuSiQue~\citep{trivedi2022musique} are multi-hop tasks: the answer usually depends on connecting more than one fact, so the first retrieval can be relevant but still incomplete. They test whether a controller knows when more evidence is needed. Natural Questions~\citep{kwiatkowski2019natural}, TriviaQA~\citep{joshi2017triviaqa}, and PopQA~\citep{mallen2023not} are open-domain factoid tasks. These are useful controls because one retrieval---or even the model's parametric knowledge---may already be enough, leaving less room for multi-step control. Every main-table cell evaluates the same $n=100$ questions for a dataset.

We use two additional diagnostic benchmarks outside the six-dataset average. RGB~\citep{chen2024rgb} replaces clean evidence with irrelevant or deliberately misleading passages, so it tests whether the controller can distinguish ``more evidence'' from ``better evidence.'' HoloBench~\citep{maekawa2025holobench} asks for sets of database rows rather than a single fact, so it tests whether a retrieval strategy can recover enough distinct items without reading the whole candidate pool. These diagnostic numbers are reported separately because their metrics are not directly comparable with QA accuracy.

\subsection{Baselines and Backbones}

The primary comparison uses \textit{gpt-oss-120b}~\citep{openai2025gptoss}. We also repeat the six-dataset evaluation with \textit{Qwen3-32B}~\citep{yang2025qwen3} as a second backbone; those results are reported in Appendix Tables~\ref{tab:qwen_answer_metrics}--\ref{tab:qwen_cost_recall}. Within each backbone, the model writes search queries and final answers, while the shared harness executes retrieval and every controller action. The Qwen study uses the same HotpotQA train/dev calibration protocol, refitted for that backbone and then frozen across all six evaluation datasets. The baselines isolate increasingly adaptive forms of retrieval control. \textbf{No-RAG} is the closed-book control~\citep{brown2020language}. \textbf{Static RAG} retrieves once and then answers~\citep{lewis2020retrieval}. \textbf{Iterative RAG} always follows the fixed multi-round schedule, showing what brute-force extra search can buy~\citep{jiang2023active}. \textbf{Adaptive-$k$} retrieves once but varies how many passages are kept from the score distribution~\citep{taguchi2025adaptivek}, while \textbf{Adaptive-RAG} routes each question once to no, single-step, or iterative retrieval based on predicted complexity~\citep{jeong2024adaptiverag}. \textbf{Self-RAG*} is a common-harness retrieval-trigger variant that isolates the decision to retrieve again~\citep{asai2024selfrag}; \textbf{CRAG} evaluates the current evidence and, when it is poor, prunes, rewrites, and retrieves again~\citep{yan2024corrective}. \textbf{RL-Search} is a prompted multi-step search controller using the Search-R1/R3-RAG interaction format, not an RL-trained reproduction~\citep{jin2025searchr1,li2025r3rag}. \method uses the evolving evidence state to decide whether to correct, retrieve again, rewrite, answer, stop, or abstain.

Unless an analysis states otherwise, every retrieving method can make at most three retrieval calls, receives top-$k=5$ passages per call, can take at most six controller actions, and has a 12k-token budget. These limits are separate: retrieval rounds control how often search can occur, top-$k$ controls how many passages one search returns, the action limit caps controller decisions, and the token budget measures the total text processed by the model.

\subsection{Metrics and Evaluation Protocol}

The primary QA metric for the benchmark and cross-backbone comparisons is normalized token-level answer F1, a deterministic overlap score already recorded by the harness. We use F1 for these headline comparisons because it does not depend on an LLM judge. Exact match (EM) is a stricter secondary answer metric, while LLM-judged semantic accuracy is reported as a supplementary semantic check. The judge shares the evaluated backbone, so 	\texttt{acc\_judge} is interpreted within a backbone and is not used for cross-backbone comparisons; headline transfer claims use deterministic token F1. Controlled mechanism and robustness analyses retain semantic ACC where their interventions were originally evaluated with that metric, and label it explicitly. Efficiency is measured by mean total tokens per question,
\begin{equation}
\mathrm{Tokens}=\frac{1}{N}\sum_{i=1}^{N} T_i,
\label{eq:metric_tokens}
\end{equation}
where $T_i$ includes retrieved text each time it is sent to the model, verifier calls, query rewriting, policy prompts, and final answer generation. We also report retrieval rounds, action counts, and evidence recall when supporting-fact annotations exist. Evidence recall is the fraction of annotated supporting facts recovered in the retained evidence.

For HoloBench, \emph{row recall} is the fraction of gold rows recovered, while \emph{recall per row read} divides row recall by the mean number of rows inspected (scaled by $10^3$ in the figure). The latter measures retrieval efficiency rather than answer quality. The complementary HoloBench result is reported in Appendix Figure~\ref{fig:holobench}, separate from the six-dataset QA comparison.

For signal analyses, AUC measures how well a score ranks positive states above negative ones, while expected calibration error (ECE) measures how closely predicted probabilities match empirical frequencies; lower ECE is better. Paired comparisons always use the same questions. Confidence intervals are obtained by bootstrap resampling over questions; exact aggregation details are in Appendix~\ref{app:repro_eval}.

Calibration and evaluation are separated. For the main results, we fit a logistic answerability calibrator over the full diagnostic vector on 300 HotpotQA training examples, select the calibrator configuration on 156 HotpotQA development examples by Brier score, and then freeze it for all six datasets without per-dataset retuning. Dedicated causal analyses use disjoint \texttt{xfit}/\texttt{xsel}/\texttt{xhold} partitions so that fitted mappings, operating choices, and final evaluation remain separated; Appendix~\ref{app:repro_data} gives the exact protocols and contamination checks.

%% file: sections/5_result.tex
\section{Results}
\label{sec:results}

\begin{table*}[!t]
\centering
\caption{
\textbf{Main comparison on six QA benchmarks with gpt-oss-120b~\citep{openai2025gptoss}
as the language backbone.}
The primary metric is normalized token-level answer F1; all cells use $n=100$ questions. Mean is the simple average across the six datasets and Tokens is mean total token cost per question. Cell color shows the F1 change from Static RAG on the same dataset. Bold marks the best F1 in each column.
}
\label{tab:main_results}
\small
\setlength{\tabcolsep}{1.2pt}
\renewcommand{\arraystretch}{1.12}
\begin{tabular*}{\textwidth}{@{\extracolsep{\fill}}l ccc ccc cc@{}}
\toprule
\multirow{2}{*}{\textbf{Method}}
& \multicolumn{3}{c}{\textbf{Multi-hop QA}}
& \multicolumn{3}{c}{\textbf{Open-domain QA}}
& \multirow{2}{*}{\textbf{Mean}}
& \multirow{2}{*}{\textbf{Tokens}} \\
\cmidrule(lr){2-4}\cmidrule(lr){5-7}
& \textbf{HotpotQA} & \textbf{2Wiki} & \textbf{MuSiQue}
& \textbf{NQ} & \textbf{TriviaQA} & \textbf{PopQA} & & \\
\midrule
No-RAG~\citep{brown2020language}
& \lossthree{.378} & \lossthree{.411} & \lossthree{.167}
& \losstwo{.341} & \gaintwo{\textbf{.766}} & \gaintwo{.381} & \lossthree{.407} & 0.10k \\
Static RAG~\citep{lewis2020retrieval}
& .559 & .681 & .423 & .413 & .681 & .324 & .514 & 2.82k \\
Adaptive-$k$~\citep{taguchi2025adaptivek}
& \lossone{.535} & \lossthree{.558} & \losstwo{.360}
& \losstwo{.347} & \gainone{.715} & \gainone{.348} & \lossone{.477} & 3.04k \\
Adaptive-RAG~\citep{jeong2024adaptiverag}
& \gainone{.575} & \gainone{.711} & \gainone{.445}
& \gainone{.425} & \gaintwo{.751} & .324 & \gainone{.538} & 3.59k \\
Iterative RAG~\citep{jiang2023active}
& \gainone{.588} & \gainone{.710} & .420
& .420 & \gaintwo{.750} & \gainthree{.440} & \gainone{.555} & 6.35k \\
CRAG~\citep{yan2024corrective}
& \gainone{.573} & \gainone{.725} & \gainone{.433}
& \gainone{.423} & \gainone{.709} & \gainthree{\textbf{.454}} & \gainone{.553} & 4.97k \\
Self-RAG*~\citep{asai2024selfrag}
& \losstwo{.482} & \lossone{.654} & \losstwo{.356}
& \lossthree{.246} & \lossthree{.100} & \lossone{.282} & \lossthree{.353} & 3.17k \\
RL-Search~\citep{jin2025searchr1,li2025r3rag}
& \lossthree{.303} & \lossthree{.308} & \lossthree{.223}
& \lossthree{.102} & \lossthree{.275} & \losstwo{.230} & \lossthree{.240} & 3.17k \\
\textbf{\method}
& \gainone{\textbf{.602}} & \gaintwo{\textbf{.766}} & \gaintwo{\textbf{.495}}
& \gainone{\textbf{.441}} & \gainone{.716} & \gaintwo{.410} & \gaintwo{\textbf{.572}} & 3.89k \\
\bottomrule
\end{tabular*}
\parbox{0.985\textwidth}{\scriptsize\color{black!70}
\textit{Color bins use absolute F1 change from Static RAG: light $[.01,.05)$, medium $[.05,.10)$, dark $\geq .10$; symmetric bins are used for losses. All rows use the same gpt-oss-120b evaluation questions; No-RAG, Adaptive-$k$, and Adaptive-RAG are from the same evaluation run as the other rows. Self-RAG* and RL-Search are common-harness mechanism variants rather than full reproductions; see Section~\ref{sec:experiments}.}}
\end{table*}

\subsection{Main Result: Quality and Cost}

\begin{figure}[!t]
    \centering
    \includegraphics[width=0.94\linewidth]{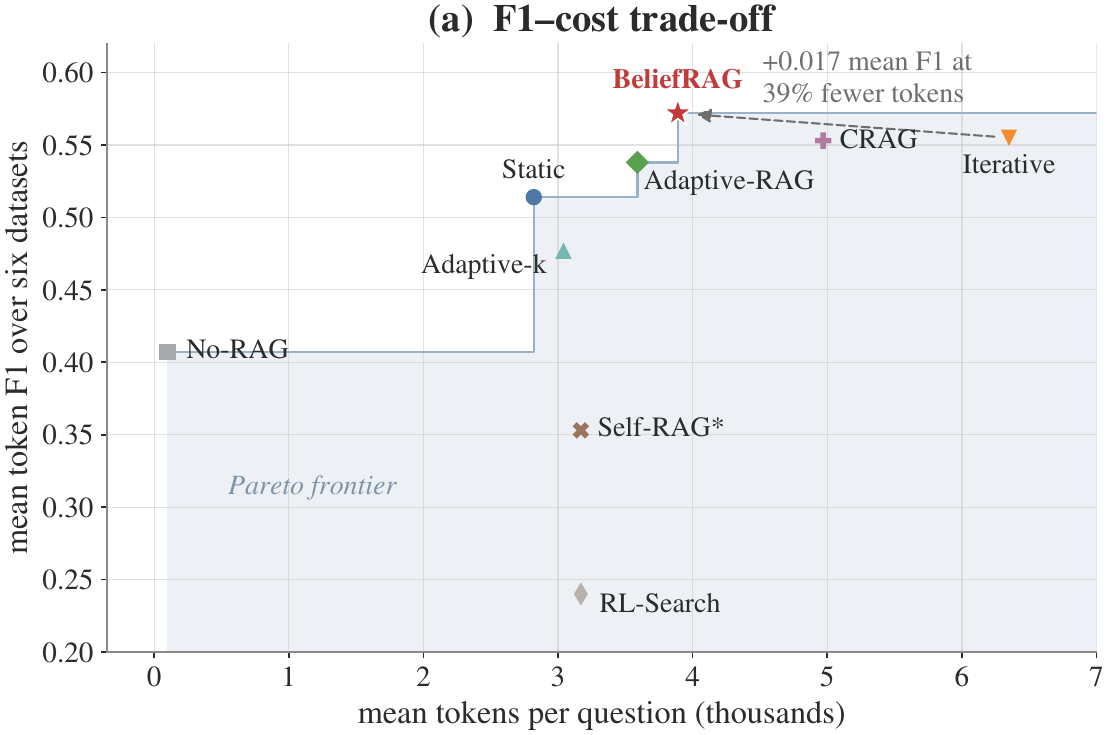}
    \caption{\textbf{F1--cost trade-off.} Each point is a method average over the six QA datasets in Table~\ref{tab:main_results}.}
    \label{fig:quality_cost}
\end{figure}

Table~\ref{tab:main_results} compares all nine methods under the same gpt-oss-120b generator, retriever, verifier, prompt, and evaluation set. Figure~\ref{fig:quality_cost} plots the same comparison against token use.

\textbf{\method achieves the highest mean F1 while using substantially less retrieval compute.}
\method reaches mean F1 $0.572$ versus $0.555$ for fixed Iterative RAG, while using $3.89$k rather than $6.35$k tokens per question---about $39\%$ less. \method is higher on four of six datasets (HotpotQA, 2Wiki, MuSiQue, and NQ), while Iterative RAG is higher on TriviaQA and PopQA. The result therefore reflects a quality--cost improvement rather than uniform gains on every dataset.

\textbf{The same pattern is stronger with a second backbone.}
With Qwen3-32B, \method reaches $0.552$ mean F1 versus $0.523$ for Iterative RAG while using $3.78$k versus $5.81$k tokens, a $35\%$ reduction. It is higher on five of six datasets; full F1, EM, LLM-judge, token, and retrieval results are reported in Appendix Tables~\ref{tab:qwen_answer_metrics}--\ref{tab:qwen_cost_recall}.

\subsection{Analysis 1: Which Parts Create the Gain?}

This analysis separates two roles of the controller. Panel (a) asks whether detecting weak evidence is enough; Panels (b--c) intervene on calibrated answerability $p_t^{\mathrm{ans}}$ (legacy run label: ``sufficiency''), not on $b_t^{\mathrm{suff}}$.

\begin{figure*}[!t]
    \centering
    \includegraphics[width=0.90\textwidth]{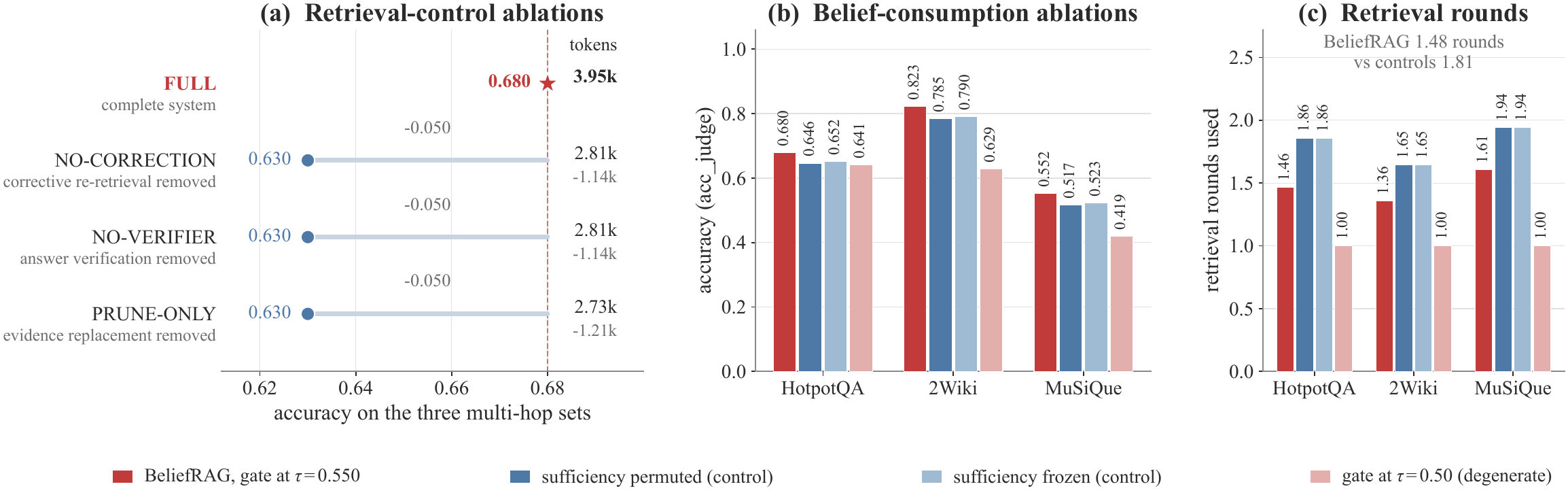}
    \caption{\textbf{Mechanism ablations.} (a) Retrieval-control ablations. (b--c) Interventions on $p_t^{\mathrm{ans}}$; ``sufficiency'' is the legacy run label.}
    \label{fig:mechanism}
\end{figure*}

\textbf{The gain comes from replacing weak evidence, not just detecting it.}
Figure~\ref{fig:mechanism}(a) shows that the full controller reaches $0.680$ multi-hop ACC. Removing correction, removing verification, or pruning weak evidence without replacing it all reduce ACC to $0.630$. This means that identifying a bad passage is not enough. The controller improves only when it removes weak evidence and retrieves a replacement that fills the missing information.

\textbf{The live answerability signal helps the controller stop once the current state becomes answerable.}
With the answer threshold fixed at $\tau=0.550$, live $p_t^{\mathrm{ans}}$ reaches $0.685$ ACC across the three multi-hop datasets. Permuting that score across matched states lowers ACC to $0.650$, while freezing it throughout the trajectory gives $0.655$. The live score also uses fewer retrieval rounds: $1.48$ on average, compared with $1.81$ for both controls. This intervention isolates the calibrated answer gate; it does not manipulate $b_t^{\mathrm{suff}}$.

\keytakeaway{The gain comes from replacing weak evidence, not merely detecting it. Calibrated answerability then tells the controller when to stop searching and answer.}

\subsection{Analysis 2: Do Thresholds Keep the Same Meaning?}

The controller answers when a score crosses a threshold, so transfer requires comparable score meaning across datasets. We compare raw and calibrated answerability scales and replay visited states to test gate reachability.

\begin{figure*}[!t]
    \centering
    \includegraphics[width=0.86\textwidth]{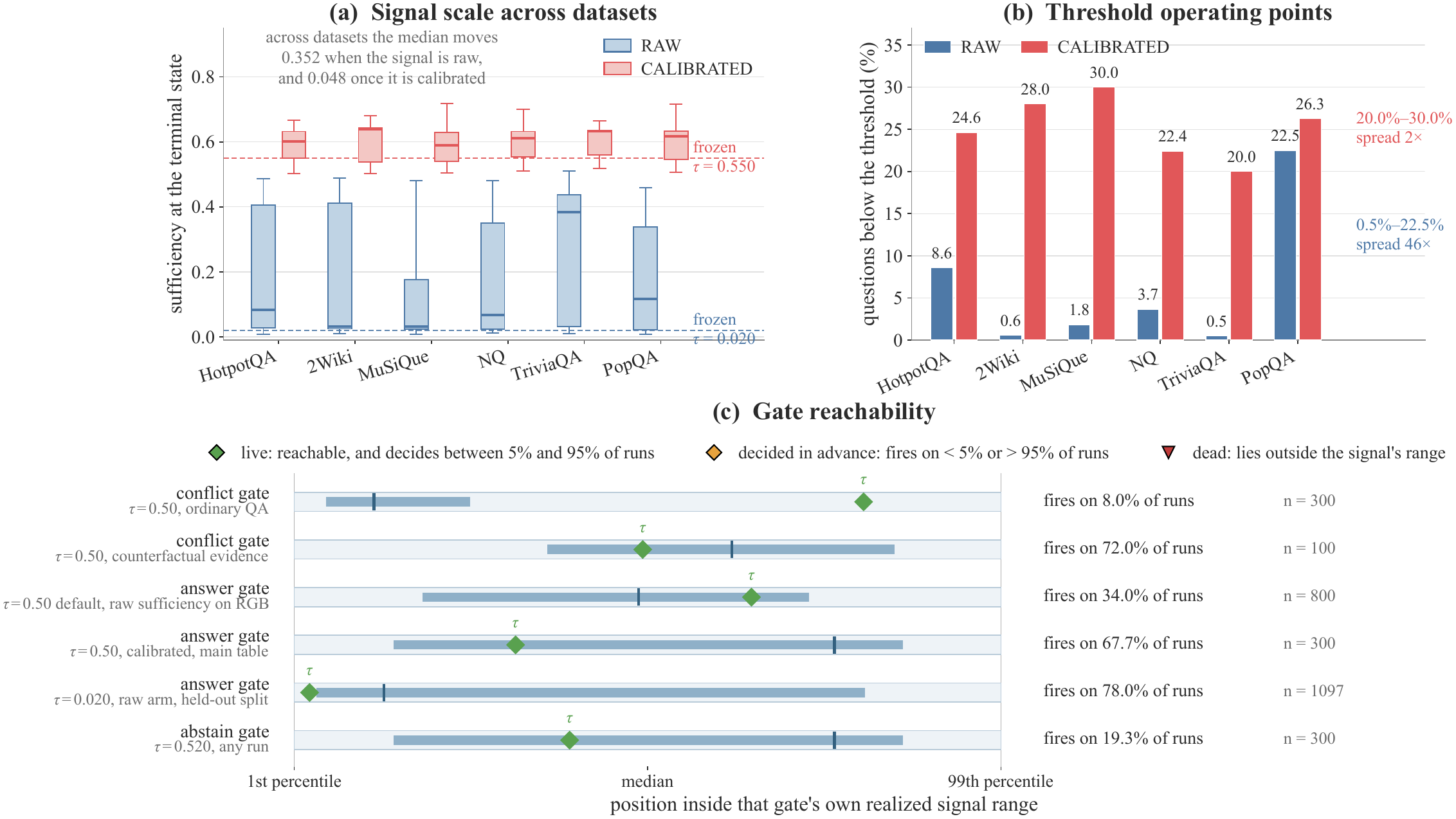}
    \caption{\textbf{Threshold transfer and reachability.} (a) Answerability scale. (b) Operating points. (c) Gate reachability. ``Sufficiency'' is the legacy label for this score.}
    \label{fig:thresholds}
\end{figure*}

\textbf{Calibration matters because thresholds depend on the scale of the score they use.}
Figure~\ref{fig:thresholds}(a) shows that the raw answerability score has substantially different medians across datasets, with a spread of $0.352$. We fit a logistic calibrator over the full diagnostic vector on HotpotQA train, select it on HotpotQA development data by Brier score, and then freeze it across all six datasets without retuning. Calibration reduces the median spread to $0.048$. Panel (b) shows the consequence: the raw threshold occupies very different parts of the score distribution across datasets, whereas the calibrated threshold lies in a more consistent operating region.

\textbf{A configured gate matters only if real trajectories can reach it.}
Panel (c) measures how often gate conditions are satisfied on visited states. The calibrated answer gate fires on $67.7\%$ of audited main-table states. The conflict gate fires on $72\%$ of counterfactual-evidence cases but only $8\%$ of ordinary QA. This is the intended behavior of a specialized signal: it remains quiet when the failure is absent and becomes active when that failure appears. A gate that stays near $0\%$ or $100\%$ across inputs would contribute little adaptive behavior.

\keytakeaway{A fixed threshold is useful only if its score keeps the same meaning. Calibration stabilizes that meaning, while reachability checks whether the gate actually affects visited states.}

\subsection{Analysis 3: What Information Is in the Belief State?}

\begin{figure*}[!t]
    \centering
    \includegraphics[width=0.87\textwidth]{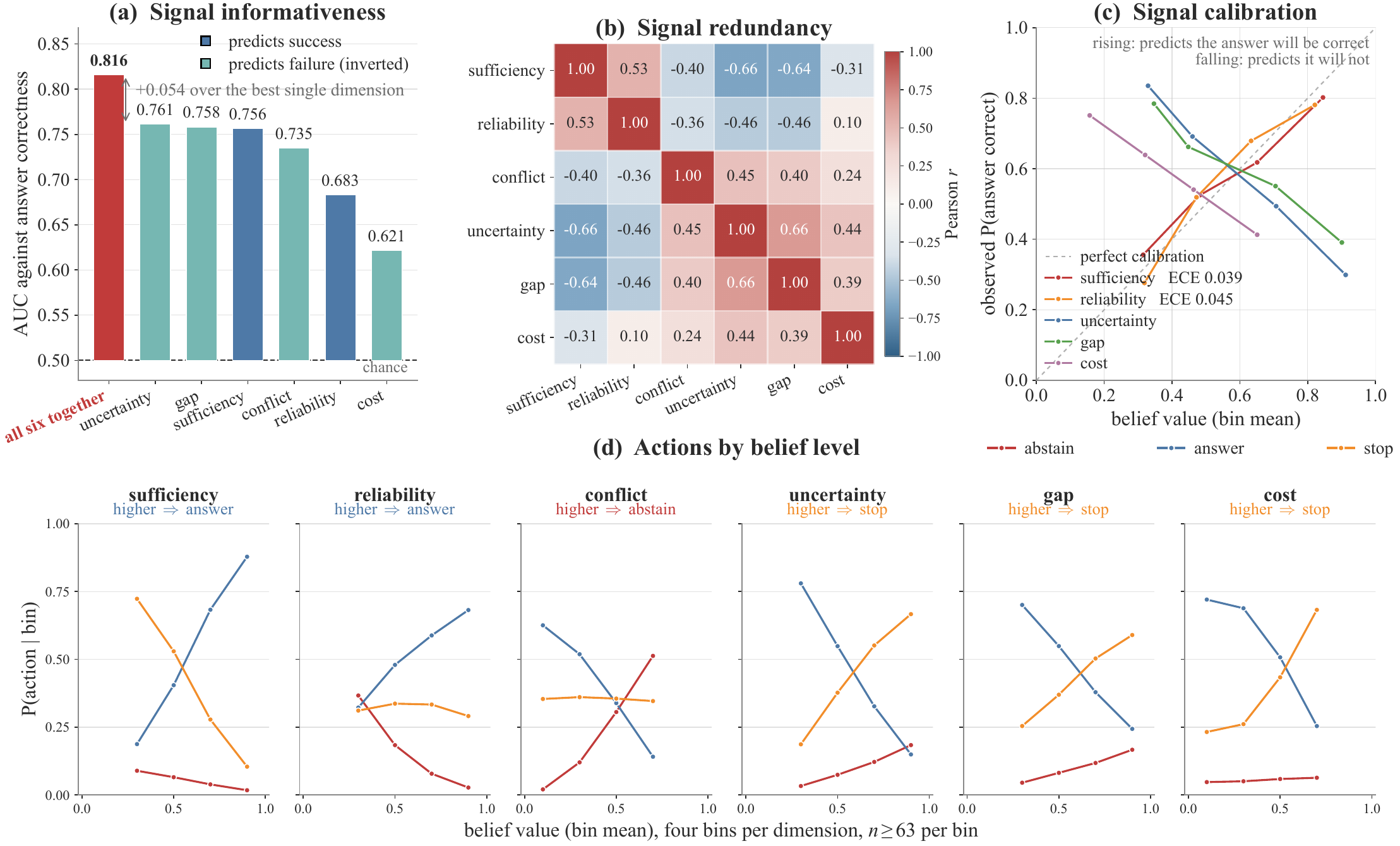}
    \caption{\textbf{Belief-state diagnostics.} (a) Signal usefulness. (b) Signal dependence. (c) Signal calibration. (d) Actions by belief level.}
    \label{fig:belief_diagnostics}
\end{figure*}

The six belief dimensions summarize different aspects of the same evidence set. We therefore ask both whether each dimension predicts correctness and whether it contributes information beyond the others. Figure~\ref{fig:belief_diagnostics} combines single-signal prediction, dependence, calibration, and action patterns across belief levels.

\textbf{The joint state is more informative than any single dimension, but the signals are correlated and unevenly useful.}
Panel (a) measures how well each signal ranks correct versus incorrect states. The joint state reaches AUC $0.816$, compared with $0.761$ for the strongest single signal, uncertainty. Uncertainty, gap, and sufficiency are the strongest individual predictors; conflict is also informative, while reliability and cost are weaker. Panel (b) shows that several dimensions are correlated because they reflect shared problems such as missing evidence. Their value therefore comes from complementary information and downstream use, not from signal count alone.

\textbf{Useful belief scores should be calibrated and linked to different controller behavior.}
Panel (c) shows that sufficiency and reliability are well calibrated, with ECE values of $0.039$ and $0.045$. Panel (d) shows the corresponding behavioral association: higher sufficiency and reliability are followed more often by answering, whereas higher uncertainty, gap, and cost are associated with stopping or further retrieval. These results establish interpretation and association for the belief dimensions. The intervention in Figure~\ref{fig:mechanism} instead tests the separate calibrated answerability score $p_t^{\mathrm{ans}}$; it should not be read as a causal intervention on $b_t^{\mathrm{suff}}$.

\keytakeaway{The value of a belief signal comes from added information, calibrated meaning, and a useful effect on decisions--not from the number of state dimensions.}

\begin{figure*}[!t]
    \centering
    \includegraphics[width=0.90\textwidth]{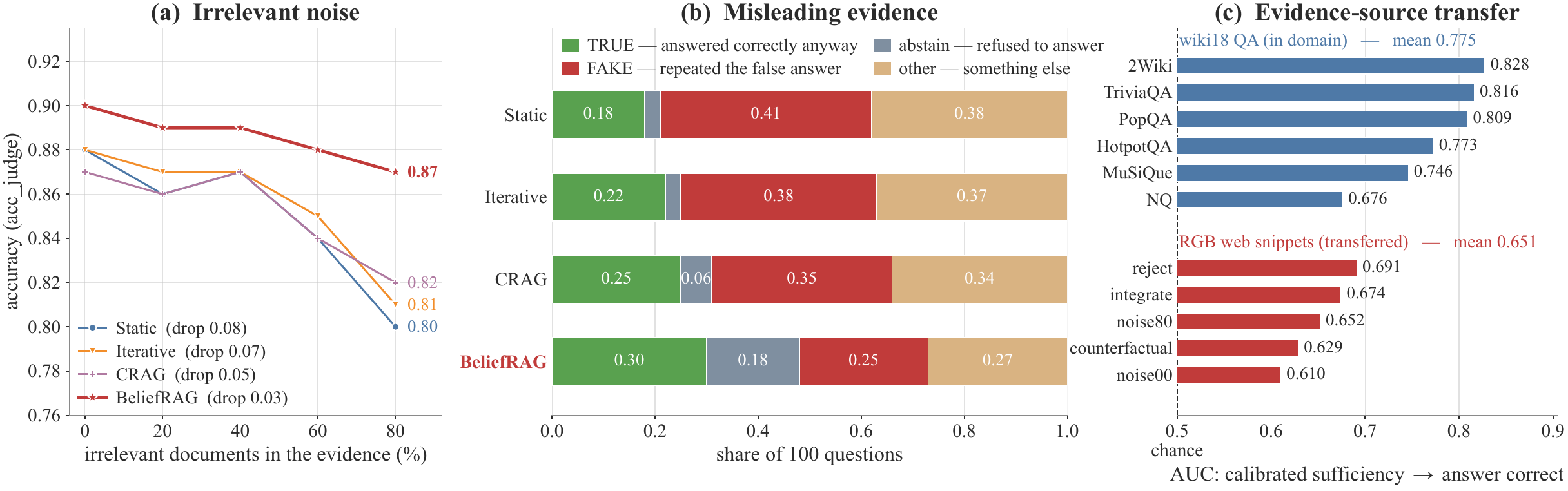}
    \caption{\textbf{Robustness to imperfect evidence.} (a) Irrelevant noise. (b) Misleading evidence. (c) Evidence-source transfer.}
    \label{fig:robustness}
\end{figure*}

\subsection{Analysis 4: What Happens When Evidence Is Imperfect?}

RGB separates three evidence failures often conflated in QA. Irrelevant noise adds useless but non-false passages; counterfactual evidence actively supports a wrong answer; source transfer tests whether the same answerability estimator still works when evidence comes from a different source. These settings probe distinct failure modes and should not be reduced to one ``robustness'' score.

\textbf{Ignoring irrelevant evidence is easier than resisting plausible but false evidence.}
Figure~\ref{fig:robustness}(a) replaces retrieved passages with irrelevant documents. At $80\%$ injection, \method retains $0.87$ ACC, versus $0.80$ for Static, $0.81$ for Iterative, and $0.82$ for CRAG, only three points below clean performance. Panel (b) is harder: \method follows the false answer on $25\%$ of questions, versus $35$--$41\%$ for the baselines, while recovering the true answer on $30\%$. Robustness to irrelevant text therefore does not imply robustness to coherent misinformation.

\textbf{Calibration can align score scales, but not guarantee source invariance.}
Panel (c) applies the same answerability calibrator to Wikipedia QA evidence and RGB web snippets. Average AUC drops from $0.775$ to $0.651$. Because AUC measures ranking rather than threshold placement, the drop reflects a less predictive evidence representation under source shift, not only threshold miscalibration. Calibration can stabilize related datasets but cannot guarantee transfer across qualitatively different evidence sources.

\keytakeaway{Robustness depends on the failure type. Filtering irrelevant or misleading passages does not solve source shift, where the diagnostic itself can lose ranking quality.}

%% file: sections/6_Analysis_Conclusion.tex
\section{Conclusion and Limitations}

\method treats retrieval as sequential evidence control. It reaches $0.572$ mean F1 with $39\%$ fewer tokens than Iterative RAG on gpt-oss-120b, and $0.552$ with $35\%$ fewer tokens on Qwen3-32B. Gains mainly come from evidence replacement and answerability-guided stopping. Limitations include lexical retrieval, only $n=100$ questions per dataset, backbone-specific calibration, self-judged auxiliary metrics, and degraded transfer under evidence-source shift.

%% file: sections/Appendix.tex
\section{Category 1: Implementation Details}
\label{app:implementation}

This appendix specifies the implementation behind the controller in the main
text. We use \emph{belief state} operationally: $b_t$ is a persistent estimate
of the current evidence condition, computed from observable diagnostics and the
previous state. The deployed system does not perform an exact Bayesian update
over an explicit latent environment variable. This distinction keeps the
appendix aligned with the estimator and controller actually evaluated.

\subsection{Evidence Workspace and Diagnostics}
\label{app:evidence_diagnostics}

The retained evidence set $E_t$ is a workspace rather than an append-only
retrieval transcript. \textsc{Retrieve} can enlarge it, \textsc{Rewrite} changes
the search query without changing it, and \textsc{Verify} can shrink it by
removing passages judged unhelpful. Dropped passage identifiers are retained so
a later retrieval cannot silently reintroduce the same passage.

At the beginning of each decision step, the harness computes
\[
x_t=[R_t,S_t,C_t,U_t,G_t,N_t,K_t],
\]
where relevance $R_t$, novelty $N_t$, and normalized cost $K_t$ are computed
locally, while support $S_t$, conflict $C_t$, uncertainty $U_t$, and gap $G_t$
come from one structured verifier call. Thus the verifier observes the evidence
but does not mutate it; removal occurs only if the controller selects
\textsc{Verify}.

For retrieval relevance, the raw BM25 score~\citep{robertson2009probabilistic} $z_{t,i}$ of passage $i$ at step $t$ is first calibrated,
\[
\begin{aligned}
\tilde z_{t,i}&=\sigma\!\left(\frac{z_{t,i}-\mu_s}{\tau_s}\right),\\
\omega_{t,i}&=\frac{\exp(\tilde z_{t,i}/\tau_R)}{\sum_{j=1}^{k}\exp(\tilde z_{t,j}/\tau_R)},\\
R_t&=\sum_{i=1}^{k} \omega_{t,i}\tilde z_{t,i}.
\end{aligned}
\]
Here $k$ is the number of passages returned by one retrieval call, $\mu_s$ and $\tau_s>0$ are the retriever-score location and scale statistics, $\tau_R$ is the relevance softmax temperature, and $\sigma$ is the logistic sigmoid defined in Section~\ref{sec:method}. The default relevance temperature is $\tau_R=0.2$. Novelty is computed from the new passages and the currently retained passages,
\[
N_t = 1-\frac{1}{|\Delta E_t|}\sum_{e\in\Delta E_t}
\max_{e'\in E_{t-1}}\operatorname{sim}(e,e'),
\]
with lexical Jaccard similarity in the evaluated configuration. With no retained
passages, novelty is $1$; with no new passages, it is $0$. Normalized acquisition
cost is
\[
K_t=\min\!\left(1,\frac{\text{cumulative tokens}}{\text{token budget}}\right).
\]

\subsection{Belief Mapping and Default Parameters}
\label{app:belief_parameters}

For each non-cost dimension $d$, the instantaneous estimate uses the complete
diagnostic vector,
\[
\hat b_t^d=\sigma(\alpha_d+w_d^\top x_t),
\]
and the persistent state is updated in logit space,
\[
b_t^d=\sigma\!\left(
(1-\lambda_d)\operatorname{logit}(b_{t-1}^d)
+\lambda_d\operatorname{logit}(\hat b_t^d)
\right).
\]
The new observation therefore receives weight $\lambda_d$. Cost is observed
directly, so $b_t^{\mathrm{cost}}=K_t$ rather than being inferred.
Table~\ref{tab:belief_params} gives the default unfitted parameterization. These
coefficients are priors whose signs follow the intended meanings of the
signals. The main stopping probability uses the separately fitted answerability
calibrator described next.

\begin{table*}[t]
\centering
\scriptsize
\caption{Default belief-state parameters. Only non-zero coefficients are shown.}
\label{tab:belief_params}
\setlength{\tabcolsep}{4pt}
\renewcommand{\arraystretch}{1.08}
\begin{tabular}{@{}lccc l@{}}
\toprule
\textbf{Dimension} & $b_0$ & $\alpha$ & $\lambda$ & \textbf{Non-zero weights in $w_d^\top x_t$} \\
\midrule
Sufficiency  & 0.35 & $-1.6$ & 0.6 & $+2.6S_t,+1.2R_t,-2.2G_t,-1.0U_t,-0.6C_t$ \\
Reliability  & 0.50 & $-1.4$ & 0.5 & $+2.8R_t,+0.9S_t,-2.4C_t$ \\
Conflict     & 0.05 & $-2.2$ & 0.6 & $+4.4C_t,-0.5S_t$ \\
Uncertainty  & 0.70 & $-1.2$ & 0.6 & $+2.6U_t,+1.3G_t,-1.6S_t,+0.7C_t$ \\
Gap          & 0.90 & $-1.0$ & 0.7 & $+3.0G_t,-1.4S_t,+0.4N_t$ \\
Cost         & 0.00 & --- & 1.0 & observed directly: $b_t^{\mathrm{cost}}=K_t$ \\
\bottomrule
\end{tabular}
\end{table*}

\subsection{Main Answerability Calibrator}
\label{app:answerability_calibration}

The main-table runs use a single \texttt{logistic\_x} calibrator over the full
seven-dimensional diagnostic vector $x_t=[R_t,S_t,C_t,U_t,G_t,N_t,K_t]$:
\begin{equation}
 p_t^{\mathrm{ans}}=\sigma(\alpha+w^\top x_t).
\end{equation}
Its target is \texttt{answerable\_now}: the fixed generator receives the question and current $E_t$, produces an answer under the evaluated answer prompt, and the shared semantic-answer judge scores that output. This is an operational generator-success target, not an entailment label for $E_t$ alone. Because the prompt requires a best short answer when the documents are incomplete, the frozen backbone may use parametric knowledge in addition to retained evidence. The calibrator is fitted only on 300 HotpotQA training examples and the configuration is selected on 156 HotpotQA development examples by lowest Brier score (0.2236; development base rate 0.2501). It is then frozen and reused without retuning on 2WikiMultiHopQA,
MuSiQue, Natural Questions, TriviaQA, and PopQA. The saved artifact is marked
\texttt{contaminated: false}. Platt and isotonic mappings exist as alternative
implementations but are not used for the reported main results.

The fitted intercept is $\alpha=0.65330$. In diagnostic-vector order, the
nonzero weights are
\[
\begin{aligned}
w_{S}&=+0.53301, & w_{G}&=-0.54360,\\
w_{U}&=-0.47780, & w_{R}&=+0.08474,\\
w_{N}&=+0.06258, & w_{K}&=-0.00441.
\end{aligned}
\]
with $w_C=0.00000$. Conflict is not manually pruned: on the clean fitting data
$C_t$ is almost always zero, so it has essentially no variance and receives a
zero fitted coefficient. This also explains why conflict is evaluated
separately under counterfactual-evidence perturbations in the main analysis.

\subsection{Terminal Controller}
\label{app:terminal_policy}

The main controller is a branch policy rather than the optional LLM action
selector. It evaluates the following branches in order:
\begin{enumerate}[leftmargin=*,topsep=2pt,itemsep=1pt,parsep=0pt]
    \item finish a correction already under way;
    \item correct materially conflicting, unreliable, or explicitly flagged evidence when the corresponding correction trigger is enabled;
    \item answer when $p_t^{\mathrm{ans}}\ge\tau_{\mathrm{ans}}$ and conflict is acceptable;
    \item retrieve when the evidence is not yet answerable, budget remains, and $p_t^{\mathrm{flip}}$ clears the minimum retrieval value;
    \item rewrite before another retrieval when the previous round added little novelty; and
    \item otherwise stop, or abstain when evidence is inadequate and further acquisition has little value.
\end{enumerate}
The default thresholds are $\tau_{\mathrm{ans}}=0.50$, conflict threshold
$0.50$, reliability threshold $0.30$, minimum retrieval value $0.10$, and
low-novelty threshold $0.20$. The controlled answerability intervention in
Figure~\ref{fig:mechanism} instead uses the separately selected and then frozen
threshold $\tau=0.55$. Its historical run labels say ``sufficiency,'' but the
manipulated score is $p_t^{\mathrm{ans}}$, not $b_t^{\mathrm{suff}}$; the two
threshold values refer to different experimental configurations.

The retrieval-value model estimates
\[
p_t^{\mathrm{flip}}
=\sigma(\alpha_{\mathrm{flip}}+w_r^{\mathrm{flip}} r_t+w_N^{\mathrm{flip}} N_t+w_s^{\mathrm{flip}} b_t^{\mathrm{suff}}).
\]
When no fitted retrieval-value model is loaded, the harness uses the recorded
per-round fallback values $\{0{:}0.59,1{:}0.073,2{:}0.050,3{:}0.050\}$ rather
than inventing a score at runtime.

\section{Category 1: Worked HotpotQA Trajectory}
\label{app:worked_trace}

Table~\ref{tab:worked_trace} gives one successful trajectory because it makes two implementation
properties concrete: the evidence set can become smaller after verification,
and a later query can name a bridge entity discovered in retained evidence.
The example also exposes a limitation that aggregate answer metrics can hide.

\paragraph{Question.}
\emph{Private Music signed the drummer who was part of which English group?}
The gold answer is \emph{Beatles}.

\begin{table*}[t]
\centering
\scriptsize
\caption{Representative HotpotQA trace. $|E_t|$ is the number of retained passages after the action.}
\label{tab:worked_trace}
\setlength{\tabcolsep}{4pt}
\renewcommand{\arraystretch}{1.12}
\begin{tabularx}{\textwidth}{@{}c l X c@{}}
\toprule
\textbf{Step} & \textbf{Action} & \textbf{Decision and query/evidence change} & $|E_t|$ \\
\midrule
0 & \textsc{Retrieve} &
Initial belief is $(0.35,0.50,0.05,0.70,0.90,0.00)$. With no evidence, the structural diagnostics set uncertainty and gap to $1$. The fallback $p^{\mathrm{flip}}=0.59$ exceeds the $0.10$ retrieval threshold. Under \texttt{first\_query\_is\_question}, the original question is issued directly as the first query. Five passages are returned; only \texttt{Private\_Music} is useful. & 5 \\
1 & \textsc{Verify} &
The verifier reports approximately $R_t=0.27$, $S_t=0.30$, $G_t=0.70$, and $K_t=0.21$, with $P(\mathrm{answerable})=0.49$. Four passages are flagged unhelpful: \texttt{Myron\_\_duo\_}, \texttt{Zak\_Starkey}, \texttt{Weathermaker\_Music}, and \texttt{Oasis\_discography}. Verification prunes them, leaving only \texttt{Private\_Music}. & 1 \\
2 & \textsc{Rewrite} &
The query is reformulated as \texttt{Private Music signed drummer formerly a member of an English band}. Rewriting changes the query but performs no retrieval. & 1 \\
3 & \textsc{Retrieve} &
The evidence-aware query writer reads the retained \texttt{Private\_Music} passage, which names \emph{Ringo Starr}, and makes that bridge entity explicit: \texttt{Ringo Starr drummer member of which English group?>}. Three additional passages are returned, but all are distractors and none mentions Ringo Starr or the Beatles. & 4 \\
4 & \textsc{Answer} &
With approximately $R_t=0.34$, $S_t=0.30$, $G_t=0.60$, $K_t=0.45$, conflict $0.09$, and $P(\mathrm{answerable})=0.53$, the answer threshold $0.50$ is crossed. The final answer is \emph{The Beatles}. & 4 \\
\bottomrule
\end{tabularx}
\end{table*}

The trace makes two behaviors concrete: verification shrinks $E_t$ from five
passages to one, and the next search names the bridge entity \emph{Ringo Starr}
found in retained evidence. It also exposes a limitation. The second retrieval
does not recover a Ringo-Starr-to-Beatles passage, so the final correct answer
combines the retained clue with the backbone's frozen parametric knowledge.
This is allowed by the prompt's explicit fallback to a best short answer when
the documents are incomplete. The trace is therefore a successful control
trajectory rather than a fully retrieved two-hop proof; the logged evidence-recall
score should be read as annotation coverage, not proof that every required
supporting passage was retrieved distinctly. The trailing \texttt{>} in the
Step~3 query is reproduced from the trace and is a query-writer formatting
error, not a manuscript placeholder.

\FloatBarrier
\section{Category 1: Model-Facing Prompts}
\label{app:main_prompts}

The main terminal controller is rule based, so it does not ask an LLM to choose
the next action. Model calls are used for evidence verification, query writing
or rewriting, and final-answer generation. The first retrieval in the worked
trace uses the original question directly and therefore incurs no query-writer
call. The prompt templates below are reproduced verbatim from the evaluated
configuration. Optional controller and baseline prompts are not used by the
main BeliefRAG controller and are therefore omitted here.
Template fields such as \texttt{\{question\}} and the JSON empty array \texttt{[]} are literal prompt/schema notation, not unfinished manuscript placeholders.

\subsection{Backend System Prompt}
\label{app:prompt_system}
\begin{promptblock}
You are the language-model backend for controlled research experiments.
Follow the task instructions in the current request exactly.
Do not use external tools, external retrieval, or hidden assumptions.
Treat the evidence, state, and other context explicitly provided in the request as
the complete experimental context unless the request says otherwise.
Do not invent missing evidence or observations.
Return exactly the output format requested by the task.
If a JSON schema is requested, return valid JSON only, with no markdown fences,
commentary, or additional text.
\end{promptblock}

\subsection{Verifier Prompt (\texttt{verifier\_v1})}
\label{app:prompt_verifier}
\begin{promptblock}
You are an evidence verifier for a retrieval experiment. Judge ONLY the evidence
shown below. Do not use outside knowledge and do not retrieve anything.

Question: {question}

Evidence passages (each prefixed by its passage id):
{evidence}

Current draft answer: {draft}

Produce these judgements, each a float in [0,1]:
  "support"     : degree to which the draft answer is entailed by the evidence.
                  If there is no draft answer, judge instead how strongly the
                  evidence entails a complete answer to the question.
  "conflict"    : the strongest contradiction present, either between two
                  evidence passages or between a passage and the draft answer.
                  0.0 if the passages are mutually consistent.
  "gap"         : the fraction of the facts or sub-questions required to answer
                  the question that are still NOT supported by the evidence.
                  1.0 means nothing required is supported, 0.0 means everything is.
  "uncertainty" : how uncertain a careful reader would remain about the final
                  answer given only this evidence.
Also list "unhelpful_doc_ids": the passage ids that are off-topic, redundant, or
misleading and should be dropped from the evidence set. Use [] if none are.

Return only this JSON object:
{"support": <float>, "conflict": <float>, "gap": <float>,
 "uncertainty": <float>, "unhelpful_doc_ids": [<id>, ...]}
\end{promptblock}

\subsection{Evidence-Aware Query Writer (\texttt{query\_refiner\_v2})}
\label{app:prompt_query_refiner}
\begin{promptblock}
Question: {question}

Evidence already retrieved:
{evidence}

Search queries already issued: {prior_queries}

Identify what the question still requires that the evidence above does NOT yet
provide, and write ONE search query targeting exactly that missing piece.

Rules:
- Do not search for anything the evidence already establishes.
- If the question needs a fact about an entity the evidence has just identified,
  name that entity explicitly in the query rather than referring to it indirectly.
- Do not repeat or lightly reword a query already issued.
- If nothing further is genuinely needed, output the single word: SUFFICIENT

Output only the query text on one line, or SUFFICIENT.
\end{promptblock}

\subsection{Query Rewriter (\texttt{query\_rewrite\_v1})}
\label{app:prompt_query_rewrite}
\begin{promptblock}
Question: {question}
Current search query: {query}
Why the current query is failing: {reason}

Rewrite the search query so it retrieves better evidence. Change the wording or
target a different required fact; do not simply repeat the current query.
Output only the rewritten query, on a single line.
\end{promptblock}

\subsection{Answer Generator (\texttt{answer\_generator\_v1})}
\label{app:prompt_answer}
The evaluated prompt is reproduced verbatim. Its first sentence is evidence-first, but the explicit fallback requires a best short answer when the documents are incomplete; parametric fallback is therefore allowed in the operational answerability target.
\begin{promptblock}
Answer the question using only the documents below. Give only the final answer,
as short as possible, with no explanation and no restatement of the question.
If the documents do not contain the answer, reply with your single best short
answer anyway.

Documents:
{evidence}

The question: {question}

Output the bare answer text only: no citations, no document numbers, no markup,
no quotation marks, and no leading "Answer:".
\end{promptblock}

\section{Category 1: Controlled Harness and Reproducibility}
\label{app:repro_harness}

This section records the implementation details needed to reproduce the
controlled comparison: execution invariants, budgets, baseline adaptations,
model settings, data partitions, aggregation, and replay safeguards.

\subsection{Execution Invariants}
\label{app:repro_invariants}

Each episode follows the same four-stage loop: (1) the verifier observes the
current evidence and returns diagnostics $x_t$; (2) the updater maps $x_t$ and
the previous belief into $b_t$; (3) the controller proposes one legal action;
and (4) the harness executes that action. The backend model does not retrieve
and does not execute actions. Only the evidence manager may mutate $E_t$, and
every mutation is logged. The feasible action set is enforced by the harness
rather than trusted to a model response.

The controller may observe the question, retained evidence, current belief,
current diagnostics, prior actions and queries, remaining budget, and the
feasible action set. Evaluation-only fields such as gold answers, supporting
facts, labels, and metrics are denied to model-facing state projections.

\subsection{Shared Budgets and Cost Accounting}
\label{app:repro_budgets_sec}

\begin{table}[t]
\centering
\small
\caption{Default budgets used by the controlled harness.}
\label{tab:repro_budgets}
\begin{tabular}{lr}
\toprule
Budget & Value \\
\midrule
Maximum retrieval rounds & 3 \\
Passages per retrieval (top-$k$) & 5 \\
Maximum controller actions & 6 \\
Token budget & 12,000 \\
Answer evidence window & 10 passages \\
\bottomrule
\end{tabular}
\end{table}

Table~\ref{tab:repro_budgets} gives the shared acquisition limits. Token cost counts retrieval context each time it is model-facing, rewriting, verification, answer generation, and controller prompting; retrieval calls are logged separately. Input/output tokens are stored separately, using a recorded local tokenizer when the API omits counts.

\subsection{Baseline Implementations}
\label{app:repro_baselines}

The common harness isolates controller mechanisms: Static retrieves once; Iterative follows fixed rounds; Adaptive-$k$ changes retained context size; Adaptive-RAG routes once by complexity; Self-RAG* isolates a retrieve/no-retrieve trigger; CRAG uses prune--rewrite--re-retrieve; RL-Search uses a prompted search/answer interface; and No-RAG is closed-book. Unless stated otherwise, these are common-harness mechanism variants rather than exact end-to-end reproductions.

\subsection{Model Backend}
\label{app:repro_backend}

Both studies use a university-hosted inference API with transmitted identifiers \texttt{gpt-oss:120b}~\citep{openai2025gptoss} and \texttt{qwen3:32b}~\citep{yang2025qwen3}. The API does not disclose the Qwen checkpoint/revision, quantization, or serving stack. Calls set temperature $0$ and \texttt{stream=false}; no \texttt{top\_p}, \texttt{top\_k}, repetition penalty, seed, stop sequence, or \texttt{max\_tokens} is transmitted. Local output limits are therefore accounting targets. A direct server probe stopped near 863 output tokens, while typical experiment outputs are about 40.

Temperature $0$ is not deterministic on this service: one fixed Qwen prompt produced two outputs over eight repeats (7/8 and 1/8), and a prior gpt-oss cache audit found 58 divergent keys among 5,341 repeats. Cached/resumed runs preserve observed responses, but uncached reruns may differ. Context probes succeeded near 4k tokens and returned HTTP 200 with an empty message near 8k and above; normal inputs are about 600 tokens, with at most 10 answer passages and 8 verifier passages, and empty generations are retried once. Within a backbone, generator, verifier, query writer, and semantic judge share one backend, so \texttt{acc\_judge} is self-judged and not used as a cross-backbone scale. Platform-returned retrieval contexts are discarded.

\subsection{Data Partitions}
\label{app:repro_data}

For each backbone, 300 HotpotQA training examples fit the answerability calibrator and 156 development examples select it by Brier score; the retrieval-value model is also refit. Qwen evaluation exits if either Qwen-specific artifact is missing rather than reusing gpt-oss artifacts. Dedicated causal analyses use deterministic \texttt{xfit}/\texttt{xsel}/\texttt{xhold} splits of 40/30/30\%, assigned by salted question-id hash, with holdout labels excluded from fitting. Within each dataset, methods share corpus, retriever, generator, verifier, answer prompt, evaluator, and budget; run metadata records the corresponding configurations.

\subsection{Metrics and Aggregation}
\label{app:repro_eval}

Per episode the harness stores EM, normalized token F1 (primary), \texttt{acc\_cover}, supplementary \texttt{acc\_judge}, evidence recall when annotated, answer/abstain status, retrieval rounds, actions, calls, tokens, and latency. Abstention scores zero on answer metrics. Repeated seeds are averaged within question before bootstrap; mean intervals use 2,000 resamples and paired within-dataset comparisons use 10,000. Headline cross-dataset results report deterministic token F1 rather than the earlier LLM-judge interval.

\section{Category 2: Complementary Results}
\label{app:complementary_results}

\begin{table*}[!t]
\centering
\caption{
\textbf{Qwen3-32B answer quality.}
Each dataset cell is \textbf{F1 / EM / Judge}, where Judge is the supplementary
LLM-judged semantic accuracy. Mean is mean F1 over six datasets.
All dataset cells use $n=100$.
}
\label{tab:qwen_answer_metrics}
\scriptsize
\setlength{\tabcolsep}{2.0pt}
\renewcommand{\arraystretch}{1.08}
\begin{tabular*}{\textwidth}{@{\extracolsep{\fill}}l cccccc c@{}}
\toprule
Method & HotpotQA & 2Wiki & MuSiQue & NQ & TriviaQA & PopQA & Mean F1 \\
\midrule
No-RAG
& .265/.180/.320
& .382/.310/.420
& .154/.080/.200
& .210/.120/.280
& .512/.420/.550
& .185/.130/.220
& .285 \\

Static
& .497/.380/.560
& .643/.570/.630
& .309/.170/.360
& .365/.230/.420
& .642/.560/.680
& .284/.240/.350
& .457 \\

Iterative
& .559/.460/.610
& .659/.570/.630
& .361/.240/.410
& .412/.310/.470
& \textbf{.695}/.600/.720
& .452/.370/.510
& .523 \\

Self-RAG*
& .367/.300/.400
& .625/.517/.552
& .285/.210/.320
& .215/.110/.260
& .088/.080/.120
& .245/.210/.300
& .304 \\

CRAG
& .512/.410/.570
& .680/.590/.660
& .352/.230/.400
& .388/.250/.440
& .665/.580/.700
& .410/.340/.460
& .501 \\

RL-Search
& .278/.180/.310
& .285/.100/.310
& .198/.100/.230
& .092/.010/.120
& .248/.180/.290
& .210/.110/.250
& .219 \\

Adaptive-$k$
& .508/.390/.565
& .651/.575/.635
& .320/.180/.370
& .372/.240/.430
& .650/.565/.685
& .305/.260/.370
& .468 \\

Adaptive-RAG
& .542/.430/.590
& .668/.595/.650
& .348/.220/.390
& .395/.270/.450
& .688/.590/.710
& .438/.360/.490
& .513 \\

\textbf{\method}
& \textbf{.582}/.475/.635
& \textbf{.715}/.620/.690
& \textbf{.418}/.295/.460
& \textbf{.428}/.325/.485
& .690/.615/.715
& \textbf{.476}/.395/.530
& \textbf{.552} \\
\bottomrule
\end{tabular*}
\end{table*}

\begin{table*}[!t]
\centering
\caption{
\textbf{Qwen3-32B token cost and logged retrieval recall.}
Panel (a) reports thousands of tokens per episode.
Panel (b) reproduces the supplied retrieval-recall fields;
BeliefRAG open-domain recall values were not supplied and are shown as dashes.
The main paper uses supporting-fact evidence recall only where annotated supporting
facts are available.
}
\label{tab:qwen_cost_recall}
\scriptsize
\setlength{\tabcolsep}{2.5pt}
\renewcommand{\arraystretch}{1.06}

\textbf{(a) Tokens/episode (k)}\\[-2pt]
\begin{tabular*}{\textwidth}{@{\extracolsep{\fill}}l cccccc c@{}}
\toprule
Method & HotpotQA & 2Wiki & MuSiQue & NQ & TriviaQA & PopQA & Mean \\
\midrule
No-RAG
& .920 & 1.050 & .980 & .850 & .910 & .820 & .922 \\

Static
& 2.624 & 3.012 & 2.815 & 2.450 & 2.580 & 2.310 & 2.632 \\

Iterative
& 5.752 & 6.662 & 6.650 & 5.210 & 5.480 & 5.120 & 5.812 \\

Self-RAG*
& 3.515 & 5.193 & 4.210 & 2.850 & 1.820 & 2.650 & 3.373 \\

CRAG
& 3.820 & 4.150 & 3.950 & 3.210 & 3.420 & 3.100 & 3.608 \\

RL-Search
& 6.120 & 7.210 & 6.850 & 5.820 & 5.950 & 5.400 & 6.225 \\

Adaptive-$k$
& 3.120 & 3.450 & 3.280 & 2.890 & 2.950 & 2.720 & 3.068 \\

Adaptive-RAG
& 3.650 & 4.120 & 4.050 & 2.980 & 3.150 & 2.850 & 3.467 \\

\textbf{\method}
& 3.850 & 4.320 & 4.250 & 3.410 & 3.550 & 3.280 & \textbf{3.777} \\
\bottomrule
\end{tabular*}

\vspace{4pt}

\textbf{(b) Logged retrieval recall}\\[-2pt]
\begin{tabular*}{\textwidth}{@{\extracolsep{\fill}}l cccccc@{}}
\toprule
Method & HotpotQA & 2Wiki & MuSiQue & NQ & TriviaQA & PopQA \\
\midrule
No-RAG
& .000 & .000 & .000 & .000 & .000 & .000 \\

Static
& .790 & .978 & .817 & .760 & .880 & .720 \\

Iterative
& .798 & .990 & .888 & .810 & .910 & .850 \\

Self-RAG*
& .606 & .991 & .740 & .520 & .310 & .650 \\

CRAG
& .825 & .985 & .850 & .835 & .925 & .820 \\

RL-Search
& .510 & .620 & .580 & .340 & .520 & .480 \\

Adaptive-$k$
& .805 & .980 & .825 & .775 & .890 & .745 \\

Adaptive-RAG
& .815 & .982 & .860 & .800 & .905 & .830 \\

\textbf{\method}
& \textbf{.860}
& \textbf{.995}
& \textbf{.912}
& --
& --
& -- \\
\bottomrule
\end{tabular*}
\end{table*}

\begin{table*}[!t]
\centering
\caption{
\textbf{gpt-oss-120b secondary answer metrics.}
Each cell is \textbf{EM / Judge}; F1 is reported in the main table.
Mean gives the simple six-dataset average for each metric.
}
\label{tab:gpt_secondary_answer}
\scriptsize
\setlength{\tabcolsep}{2.1pt}
\renewcommand{\arraystretch}{1.08}
\begin{tabular*}{\textwidth}{@{\extracolsep{\fill}}l cccccc c@{}}
\toprule
Method & HotpotQA & 2Wiki & MuSiQue & NQ & TriviaQA & PopQA & Mean EM/Judge \\
\midrule
Static
& .450/.640
& .600/.720
& .290/.440
& .280/.510
& .600/.810
& .280/.330
& .417/.575 \\

Iterative
& .452/.660
& .670/.750
& .355/.520
& .350/.620
& .625/.860
& .410/.530
& .477/.657 \\

CRAG
& .460/.650
& .630/.770
& .310/.460
& .280/.570
& .630/.830
& .380/.470
& .448/.625 \\

Self-RAG*
& .380/.540
& .600/.680
& .270/.400
& .130/.310
& .100/.120
& .250/.290
& .288/.390 \\

RL-Search
& .200/.620
& .110/.750
& .120/.440
& .010/.440
& .200/.530
& .130/.420
& .128/.533 \\

\textbf{\method}
& .480/.630
& .660/.810
& .370/.520
& .290/.580
& .650/.850
& .340/.430
& .465/.637 \\
\bottomrule
\end{tabular*}
\end{table*}

\begin{table*}[!t]
\centering
\caption{
\textbf{gpt-oss-120b cost and multi-hop evidence recall.}
Tokens are thousands per episode.
Evidence recall is reported only for the three multi-hop benchmarks
with supporting-fact annotations.
}
\label{tab:gpt_cost_recall}
\scriptsize
\setlength{\tabcolsep}{2.4pt}
\renewcommand{\arraystretch}{1.06}

\textbf{(a) Tokens/episode (k)}\\[-2pt]
\begin{tabular*}{\textwidth}{@{\extracolsep{\fill}}l cccccc c@{}}
\toprule
Method & HotpotQA & 2Wiki & MuSiQue & NQ & TriviaQA & PopQA & Mean \\
\midrule
Static
& 2.620 & 2.998 & 2.835 & 2.762 & 2.834 & 2.849 & 2.816 \\

Iterative
& 5.536 & 6.616 & 6.382 & 6.738 & 6.427 & 6.425 & 6.354 \\

CRAG
& 4.395 & 5.130 & 4.946 & 5.331 & 4.599 & 5.427 & 4.971 \\

Self-RAG*
& 3.473 & 4.172 & 4.012 & 2.736 & .897 & 3.707 & 3.166 \\

RL-Search
& 3.308 & 3.757 & 3.324 & 2.985 & 2.402 & 3.251 & 3.171 \\

\textbf{\method}
& 3.472 & 4.293 & 4.081 & 4.029 & 3.364 & 4.130 & \textbf{3.895} \\
\bottomrule
\end{tabular*}

\vspace{4pt}

\textbf{(b) Evidence recall}\\[-2pt]
\begin{tabular}{lcccc}
\toprule
Method & HotpotQA & 2Wiki & MuSiQue & Mean \\
\midrule
Static
& .790 & .978 & .817 & .862 \\

Iterative
& .793 & .990 & \textbf{.898} & \textbf{.894} \\

CRAG
& .783 & .985 & .807 & .858 \\

Self-RAG*
& .691 & .927 & .742 & .787 \\

RL-Search
& .728 & .927 & .733 & .796 \\

\textbf{\method}
& .788 & \textbf{.990} & .817 & .865 \\
\bottomrule
\end{tabular}
\end{table*}

\begin{figure*}[!t]
    \centering
    \includegraphics[width=0.90\textwidth]{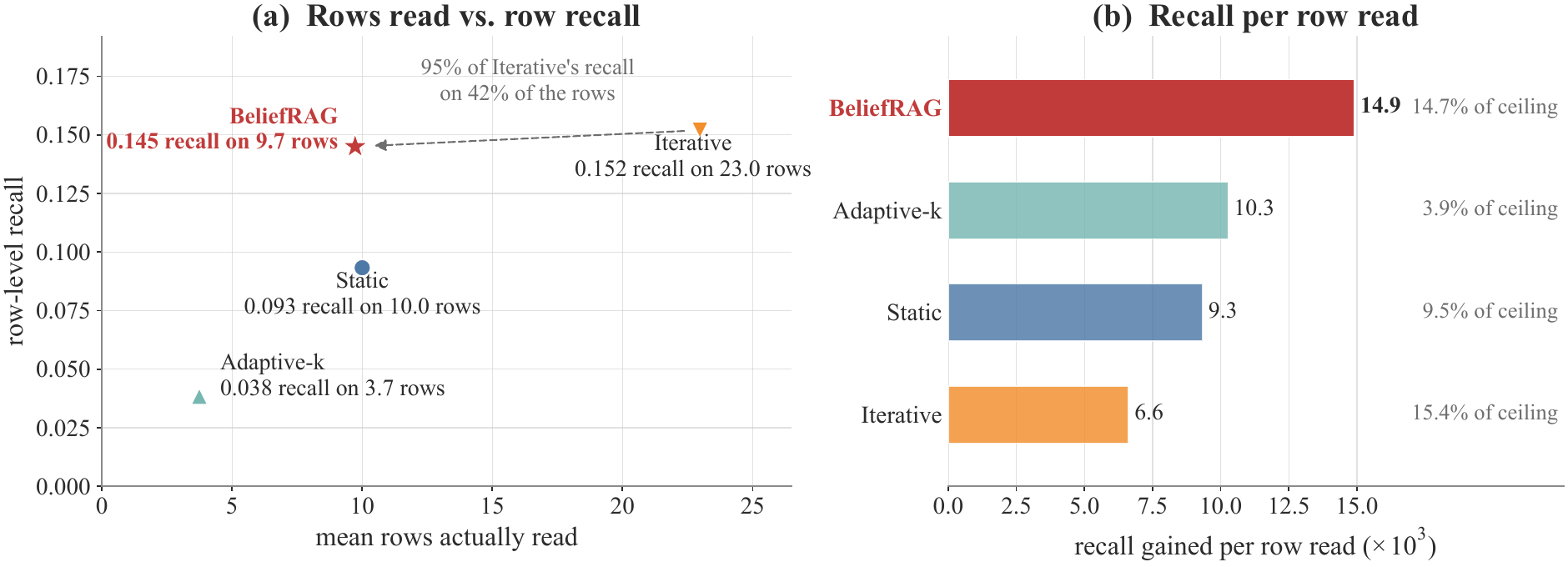}
    \caption{
    \textbf{HoloBench aggregation retrieval.}
    \emph{Row recall} is the fraction of gold rows recovered in the evidence shown to the model;
    \emph{recall per row read} divides that recall by the mean number of rows inspected
    (scaled by $10^3$ for display).
    (a) BeliefRAG reaches row recall $0.145$ after reading $9.73$ rows on average,
    versus $0.152$ after $22.98$ rows for fixed Iterative retrieval.
    (b) BeliefRAG therefore obtains higher recall per row read.
    The structural ceiling at 50 rows is $0.985$; absolute recall remains far below it,
    so this diagnostic supports a selection-efficiency claim rather than a claim that
    aggregation retrieval is solved.
    These common-harness HoloBench numbers are not directly comparable with the QA F1
    scores in Table~\ref{tab:main_results}.
    }
    \label{fig:holobench}
\end{figure*}